\documentclass[10pt,twocolumn,letterpaper]{article}

\usepackage[pagenumbers]{cvpr} 

\definecolor{cvprblue}{rgb}{0.21,0.49,0.74}
\usepackage[pagebackref,breaklinks,colorlinks,allcolors=cvprblue]{hyperref}

\def\paperID{4} 
\def\confName{CVPR}
\def\confYear{2026}

\title{DriveSafer: End-to-End Autonomous Driving with Safety Guidance}

\author{Shounak Sural\\
Carnegie Mellon University\\
Pittsburgh, PA\\
{\tt\small iamshounak@gmail.com}
\and
Ragunathan (Raj) Rajkumar\\
Carnegie Mellon University\\
Pittsburgh, PA\\
{\tt\small rajkumar@andrew.cmu.edu}
}
\begin{document}
\maketitle
\begin{abstract}
    End-to-End (E2E) autonomous driving models have shown growing capability in recent years, with performance improving on increasingly challenging benchmarks. However, modern generative E2E planners still suffer from a substantial number of catastrophic failures in safety-critical scenarios. We find that many such failures arise from violations of physical constraints and safety requirements, leading to unsafe behavior. Motivated by this finding, in this paper, we focus on improving safety outcomes in generative end-to-end driving with a targeted reduction of catastrophic planning failures, instead of enhancing average planning quality. Towards this end, we propose \textit{DriveSafer}, a failure-aware safety framework for end-to-end planners. \textit{DriveSafer} explicitly steers generative planners towards safe behaviors leveraging both training-time safety constraints and inference-time safety guidance. Compared to the state-of-the-art DiffusionDrive model, on the NAVSIM benchmark, \textit{DriveSafer} reduces the number of catastrophic failures (\texttt{PDMS=0}) by $48\%$, with over $65\%$ reduction in drivable-area compliance failures.
\end{abstract}
\vspace{-6mm}
\section{Introduction}
End-to-End (E2E) autonomous driving models have recently become increasingly popular. Compared to traditional modular and rule-based architectures, these models show promise in addressing more complex scenarios with a large number of interacting agents, and in long-tailed scenarios aided by the fusion of vision and language with internet-scale datasets. While these E2E models handle many situations well, they can also be surprisingly brittle in many commonly seen situations which are straightforward for humans and even rule-based systems to handle. These misbehaviors typically show up in the long tail in end-to-end autonomous driving benchmarks such as NuPlan \cite{caesar2021nuplan} and NAVSIM \cite{dauner2024navsim} and long-tail-focused datasets such as KIT-E2E \cite{wagner2026longtaildrivingscenariosreasoning} and WOD-E2E \cite{xu2025wod}. To identify many such failure cases, some benchmarks also employ CARLA \cite{dosovitskiy2017carla} simulation allowing closed-loop testing with benchmarks such as Bench2Drive \cite{jia2024bench2drive} and Longest6 \cite{chitta2022transfuser}. Importantly, strong average benchmark performance for models, however, still masks a long tail of catastrophic failures.

While E2E models built using popular benchmarks have improved planning over the years, even the best-performing models such as DiffusionDrive \cite{liao2025diffusiondrive} still suffer from catastrophic failures that simple rule-based planners can easily avoid. To reduce the failure modes of existing E2E models, a careful evaluation of safety is necessary. Many existing E2E models are benchmarked using the L2 distance error metric from human-driven ground truth trajectories as the primary metric on datasets such as NuScenes. While imitation learning minimizing the L2 error from ground truth offers useful insights, such metrics can be surprisingly easy to imitate \textit{without} adding actual functional value. Methods such as AD-MLP \cite{zhai2023rethinkingopenloopevaluationendtoend} and Ego-MLP \cite{li2024egostatusneedopenloop} have shown that a surprisingly simple MLP model which does \textit{not} even consume sensor inputs can outperform state-of-the-art methods on the NuScenes benchmark which optimizes only for trajectory error. Such ``blind" models that achieve state-of-the-art performance clearly underscore the need for better benchmarks and different evaluation metrics. NAVSIM \cite{dauner2024navsim} is one such benchmark that has standardized evaluation metrics which are aligned with driving safety. Their proposed Predictive Driver Model Score (PDMS) metric comprises collision checks, drivable area compliance, comfort, time-to-collision and driving direction compliance. These metrics are much better indicators for a safe planner. NAVSIMv2 \cite{cao2025pseudo} introduces the Extended PDMS metric that further incorporates  traffic-light compliance, lane-keeping and extended comfort metrics for a comprehensive safety-conscious evaluation. Earlier proposed ``blind" MLP models which can exhibit unsafe behaviors do not perform favorably under these extended evaluation metrics. 

Many modern state-of-the-art methods report their performance on NAVSIM in terms of the PDMS score, but there is less emphasis on analyzing catastrophic failure cases in these models. In this paper, we specifically consider a PDMS score of zero in the NAVSIM E2E planning benchmark as a catastrophic failure, since its occurrences can potentially be hazardous and life-threatening in real-world driving. A PDMS score of 0 is assigned in the following cases: (i) the ego vehicle collides with another moving object such as a vehicle or a pedestrian, (ii) the ego vehicle drives off the permitted drivable region possibly onto a sidewalk or a solid divider, and (iii) the ego vehicle motion differs significantly from the intended driving direction, such as going straight when explicitly directed to turn right. Thus, we take a failure-centric view of safe driving behavior and focus on reducing these catastrophic failure cases as opposed to average benchmark performance. We carefully evaluate such failure cases and propose a framework called \textit{DriveSafer} that strategically reduces these failures. Correspondingly, our contributions are as follows: 
\begin{itemize}
    \item We propose \textit{DriveSafer}, a failure-centric safety enforcement framework for E2E planning that directly targets reducing catastrophic failures using synergistic training- and inference-time safety enforcement.
    \item We introduce a safety-aware training strategy with specialized physics-informed and safety-aware losses along with lightweight and integrated forecasting and planning. 
    \item We develop an inference-time safety guidance method that rejects infeasible trajectories and steers generated trajectories towards safety-compliant ones.
    \item \textit{DriveSafer} reduces catastrophic failures (\texttt{PDMS=0}) on the NAVSIM benchmark by $48\%$ over the state-of-the-art DiffusionDrive method while also improving overall PDMS. 
\end{itemize}
\begin{figure}
    \centering
    \includegraphics[width=\linewidth]{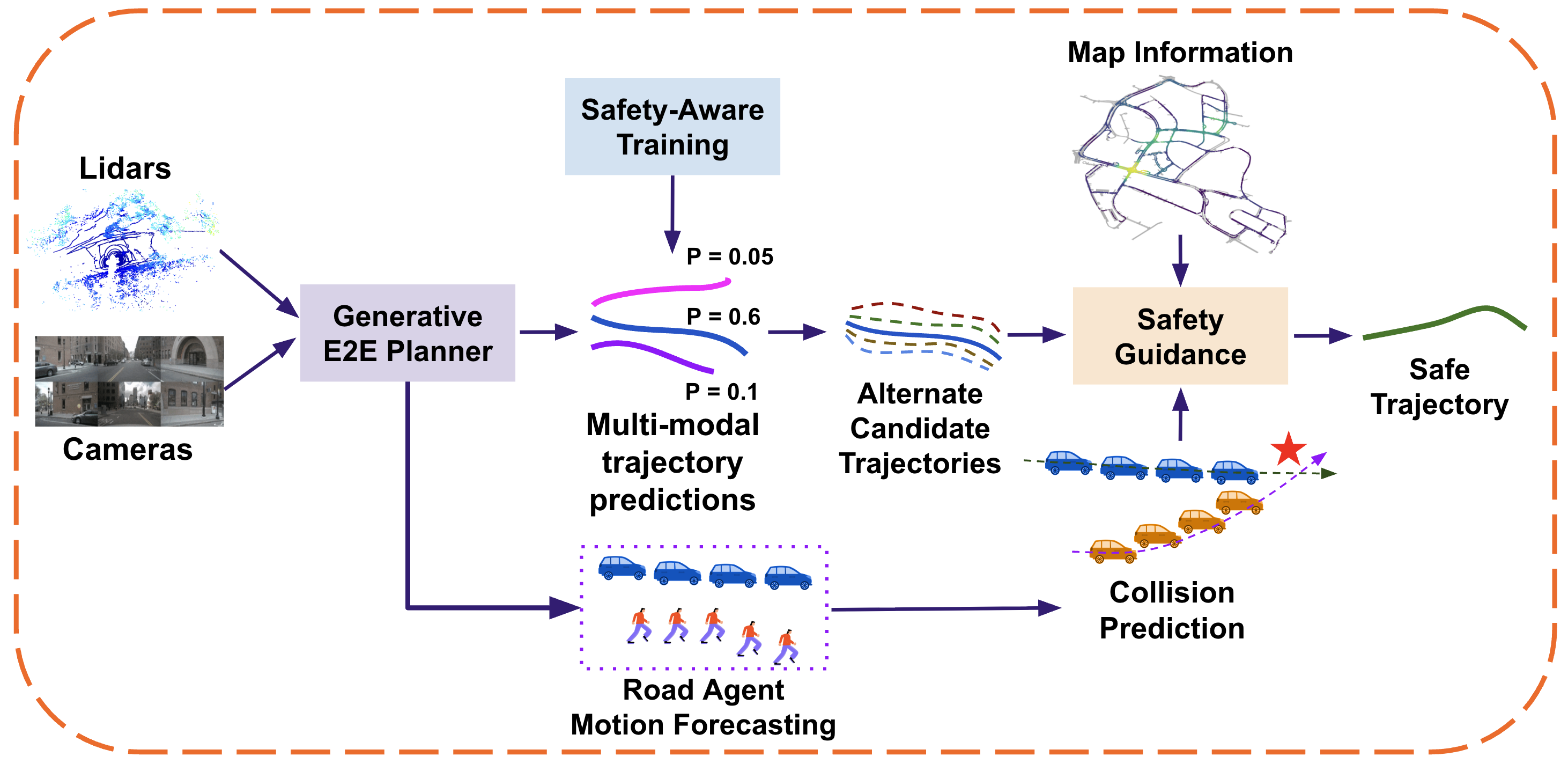}
    \caption{Overview of our \textit{DriveSafer} framework}
    \label{fig:block_diag_DriveSafer}
\end{figure}
\section{Related Work}
\textbf{End-to-End Autonomous Driving:}
Early E2E studies focused on generating driving commands such as steering angles and acceleration/brake commands directly from raw sensor inputs through imitation and reinforcement learning \cite{codevilla2018end,chen2020learning}. However, these methods do not scale across different platforms due to varying vehicle dynamics. It is now well understood that low-level actuation commands are best generated through specialized controllers for each vehicle given a local motion plan. Hence, recent end-to-end autonomous driving literature focuses on generating local motion plans from sensor data. E2E methods use strategies such as learning scene representations in the BEV space \cite{hu2023planning,chitta2022transfuser, Tang_2025_ICCV, weng2024drive}, learning vectorized architectures \cite{jiang2023vad,chen2024vadv2}, vision-language action model-based planning \cite{wang2025alpamayo,renz2025simlingo, zhang2025coc,zhou2025autovla, fu2025orion}, world model-based planners \cite{li2024think2drive, wang2024driving, zheng2024occworld, wang2024drivedreamer} and diffusion-based generative planners \cite{zheng2025diffusion,liao2025diffusiondrive}.

\textbf{Datasets for Safety in Autonomous Driving:} 
Another direction of research focuses on enforcing safety in autonomous driving via datasets and simulation that critically evaluates safety, especially in rare and long-tailed scenarios. For example, NuPlan evaluates full-scale E2E behavior using closed-loop simulation. NAVSIMv2 \cite{cao2025pseudo} uses neural reconstruction to generate alternate scenarios complete with generated sensor data for pseudo simulation. WOD-E2E \cite{xu2025wod} uses millions of miles of Waymo driving data to mine long-tailed scenarios that occur rarely in daily life ($<0.03\%$ of cases). Bench2Drive \cite{jia2024bench2drive} models challenging interactions and harsh weather conditions in CARLA \cite{dosovitskiy2017carla} for safety evaluation of models.

\textbf{Methods for Safe Autonomous Driving:} 
Another branch of work aims to make driving policies safer through methods such as CAT \cite{zhang2023cat} which turns simple real-world scenarios into alternate plausible safety-critical ones to perform closed-loop adversarial training. DriveAdapter \cite{jia2023driveadapter} injects hand-crafted safety rules into a teacher-student paradigm for training. Control Barrier Functions are sometimes used to enforce safety in planning \cite{xu2026realtimecontrolbarrierfunctionbased, allamaa2024real, konda2019provably}. LEAD \cite{nguyen2025leadminimizinglearnerexpertasymmetry} narrows the sim2real gap by minimizing unrealistic privileged information available to planning experts in the CARLA simulator \cite{dosovitskiy2017carla}.

Similar to LLM safety research, Reinforcement Learning-based alignment ideas have been applied to E2E autonomous driving with methods such as TrajHF \cite{li2025finetuning}, Carl \cite{jaeger2025carl} and DriveDPO \cite{shang2025drivedpo}. Unlike prior work that focuses on such RL-based methods, our proposed \textit{DriveSafer} framework focuses on guiding a generative planner towards safer choices by adhering to certain rules of the road. This guidance significantly reduces catastrophic failures and yields acceptable motion plans in a lot of cases. 

\begin{figure}
    \centering
    \includegraphics[width=0.95\linewidth]{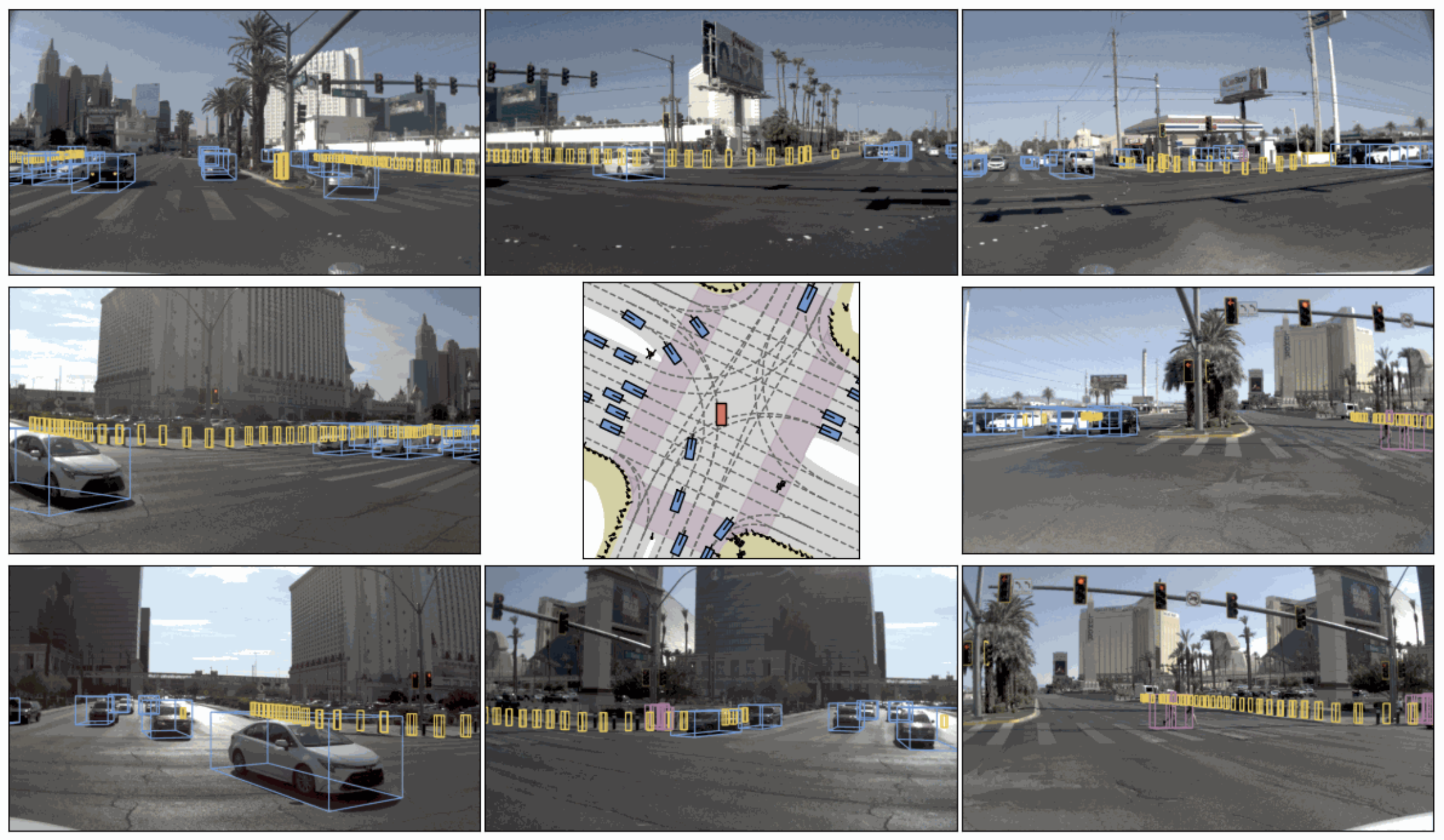}
    \vspace{-3mm}
    \caption{A complex left-turn in NAVSIM \cite{dauner2024navsim}}
    \label{fig:left_turn}
\end{figure}
\section{Methodology}
\vspace{-2mm}
An overview of our \textit{DriveSafer} framework is shown in Figure \ref{fig:block_diag_DriveSafer}. We will next describe our framework. 
\subsection{Base End-to-end Planning Model}
Our \textit{DriveSafer} framework is based on the truncated diffusion model called DiffusionDrive \cite{liao2025diffusiondrive}. We choose this primary baseline since it is a state-of-the-art method on the NAVSIM benchmark and performs inference at 45 fps, making it suitable for \textit{DriveSafer} to be deployed on a real-world AV.

\begin{figure}
    \centering
    \includegraphics[width=0.7\linewidth]{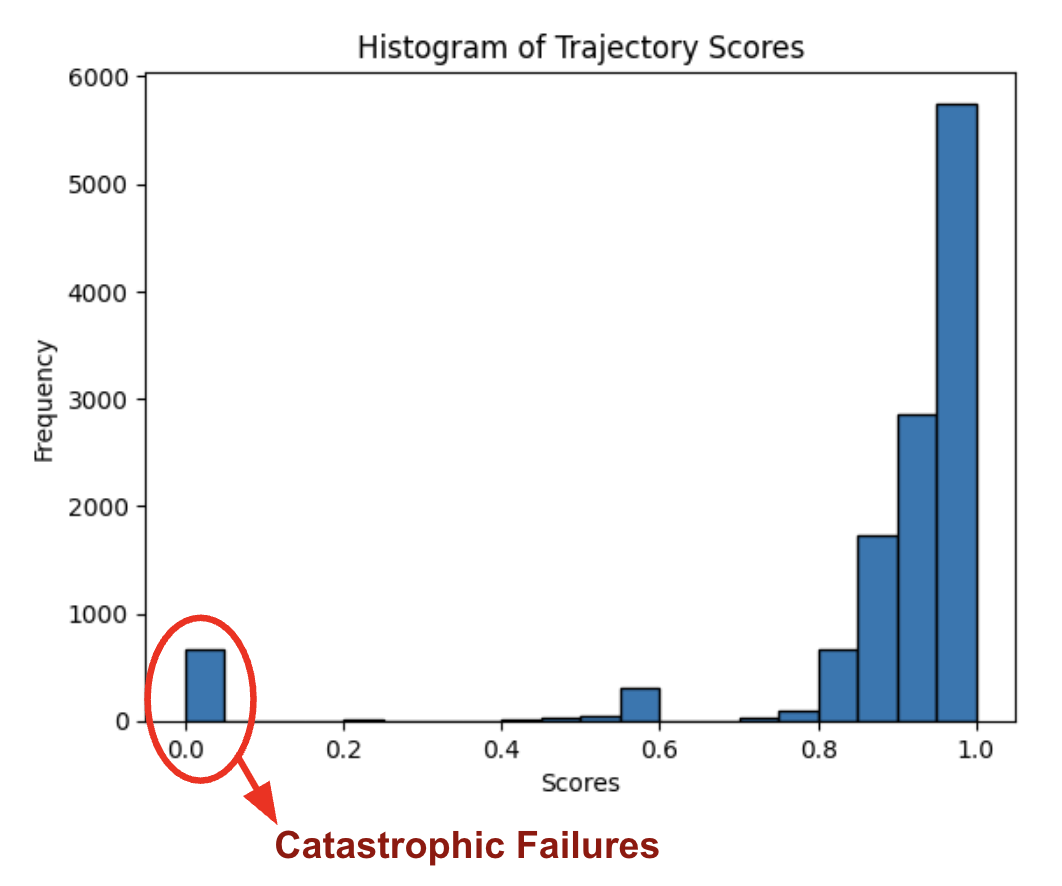}
    \vspace{-4mm}
    \caption{PDM Score Distribution for DiffusionDrive showing catastrophic failure cases with \texttt{\texttt{PDMS=0}}}
    \label{fig:diffdrive_error_dist}
\vspace{-5mm}
\end{figure}
DiffusionDrive consumes lidar and camera sensor data and learns intermediate representations necessary for object detection, tracking and map segmentation. It then trains a 2-step diffusion model conditioned on the learnt intermediate representation. Instead of sampling from pure noise like most diffusion models, DiffusionDrive starts with noise centered around a few typical trajectory modes and performs trajectory planning with mode selection followed by denoising for fine adjustments in only 2 steps. While performing well in many situations, its effectiveness for real-world use is limited due to a large number of catastrophic failures as highlighted in Figure \ref{fig:diffdrive_error_dist}. We address such drawbacks with \textit{DriveSafer}.
\vspace{-3mm}
\subsection{Safety-Aware Training in \textit{\textbf{DriveSafer}}}
Safety-aware training in \textit{DriveSafer} is intended to make the generated trajectories acceptable for driving safety. We incorporate safety-aware training in \textit{DriveSafer} using additional losses defined as follows.
\vspace{-3mm}
\begingroup
\setlength{\abovedisplayskip}{2pt}
\setlength{\belowdisplayskip}{2pt}
\setlength{\abovedisplayshortskip}{0pt}
\setlength{\belowdisplayshortskip}{2pt}
\setlength{\jot}{1pt}

\begin{equation}
L_{\text{total}} = L_{\text{base}} + \lambda_{\text{DAC}}L_{\text{DAC}} + \lambda_{\text{Col}}L_{\text{Col}} + \lambda_{\text{Comf}}L_{\text{Comf}}.
\end{equation}
\begin{equation}
L_{\text{DAC}}
= \frac{1}{BT}\sum_{b=1}^{B}\sum_{t=1}^{T}
\Big(1 - M_b(\hat{\mathbf{p}}_{b,t})\Big)
\end{equation}
\begin{equation}
d_{b,t,n} = \|\hat{\mathbf{p}}_{b,t} - \mathbf{a}_{b,t,n}\|_2
\end{equation}
\begin{equation}
\begin{split}
L_{\text{Col}}
&= \frac{1}{BT|\mathcal N_a|}
\sum_{b=1}^{B}\sum_{t=1}^{T}\sum_{n\in\mathcal N_a}
\max(0,d_{\text{col}} - d_{b,t,n})
\end{split}
\end{equation}
\begin{equation}
\mathbf{v}_{b,t} = \frac{\hat{\mathbf{p}}_{b,t+1} - \hat{\mathbf{p}}_{b,t}}{\Delta t},
\mathbf{j}_{b,t} = \frac{\mathbf{v}_{b,t+2} - 2\mathbf{v}_{b,t+1} + \mathbf{v}_{b,t}}{\Delta t^2}
\end{equation}
\begin{equation}
\begin{split}
L_{\text{Comf}}
&= \frac{1}{B(T-3)}
\sum_{b=1}^{B}\sum_{t=1}^{T-3}
\max\!\left(0,\ \|\mathbf{j}_{b,t}\|_2 - j_{\text{th}}\right)
\end{split}
\end{equation}

\endgroup
Here, $L_{\text{base}}$ is the original training loss, $B$ is the batch size denoting the number of scenes, and $T$ is the planning horizon. Among the additional loss terms, $L_{\text{DAC}}$ penalizes predicted trajectories from leaving the drivable area,
$L_{\text{Col}}$ penalizes trajectories that come too close to surrounding agents,
and $L_{\text{Comf}}$ reduces sudden motion changes to ensure comfort. Additionally, $\hat{\mathbf{p}}_{b,t}\in\mathbb{R}^2$ is the predicted ego position at time step $t$ in scene $b$,
$M_b$ is a binary drivable-area mask for scene $b$ obtained from the underlying map,
$\mathcal N_a$ is the set of valid surrounding agents
and $\mathbf{a}_{b,t,n}\in\mathbb{R}^2$ is the position of agent $n$ at time step $t$.
Furthermore, $d_{b,t,n}$ denotes the Euclidean distance between the ego and agent $n$,
$d_{\text{col}}$ is the collision-distance threshold,
$\Delta t$ is the time interval between consecutive trajectory points,
$\mathbf{v}_{b,t}$ is the ego velocity,
$\mathbf{j}_{b,t}$ is the ego jerk,
and $j_{\text{th}}$ is the jerk threshold.
 \begin{table}[t]
\centering
\caption{Comparison of \textit{DriveSafer} with previous end-to-end autonomous driving methods on the \texttt{navtest} split of NAVSIM.}
\vspace{-3mm}
\label{tab:e2e_comparison}
\setlength{\tabcolsep}{4pt}
\renewcommand{\arraystretch}{1.05}
\resizebox{\linewidth}{!}{%
\begin{tabular}{l c c c c c c c}
\toprule
\textbf{Method} & \textbf{Sensor Input} & \textbf{NC$\uparrow$} & \textbf{DAC$\uparrow$} & \textbf{TTC$\uparrow$} & \textbf{Comf.$\uparrow$} & \textbf{EP$\uparrow$} & \textbf{PDMS$\uparrow$} \\
\midrule
Constant Velocity \cite{dauner2024navsim} & - & 68.0 & 57.8 & 50.0 & \textbf{100} & 19.4 & 20.6 \\
Ego Status MLP~\cite{dauner2024navsim} & - & 93.0 &  77.3  & 83.6 & \textbf{100} & 62.8 & 65.6 \\
UniAD~\cite{hu2023planning} & Camera & 97.8 & 91.9 & 92.9 & \textbf{100} & 78.8 & 83.4 \\
LTF~\cite{chitta2022transfuser} & Camera & 97.4 & 92.8 & 92.4 & \textbf{100} & 79.0 & 83.8 \\
PARA-Drive~\cite{weng2024drive} & Camera & 97.9 & 92.4 & 93.0 & 99.8 & 79.3 & 84.0 \\
LAW~\cite{li2024enhancing} & Camera & 97.9 & 92.4 & 93.0 & 99.8 & 79.3 & 84.0 \\
DriveX \cite{Shi_2025_ICCV} & Camera & 97.5 & 94.0 & 93.0 & \textbf{100} & 79.7 & 84.5 \\
DRAMA~\cite{yuan2024drama} & C \& L & 98.0 & 93.1 & 94.8 & \textbf{100} & 80.1 & 85.5 \\
Transfuser~\cite{chitta2022transfuser} & C \& L & 97.7 & 92.8 & 92.8 & \textbf{100} & 79.2 & 84.0 \\
VADv2~\cite{chen2024vadv2} & C \& L & 97.2 & 89.1 & 91.6 & \textbf{100} & 76.0 & 80.9 \\
Hydra-MDP~\cite{li2024hydra} & C \& L & 98.3 & 96.0 & 94.6 & \textbf{100} & 78.7 & 86.5 \\
WoTE\cite{li2025end} & C \& L & \textbf{98.5} & 96.8 & \textbf{94.9} & 99.9 & 81.9 & 88.3\\
DiffusionDrive \cite{liao2025diffusiondrive} & C \& L & 98.2 & 96.2 & 94.7 & \textbf{100} & 82.2 & 88.1 \\
\midrule
\textbf{DriveSafer (Ours)} & \textbf{C \& L} & 98.3 & \textbf{98.7} & 94.6 & \textbf{100} & \textbf{84.4} & \textbf{90.3} \\

\bottomrule
\end{tabular}%
}
\end{table}

\begin{figure*}
    \centering
    \includegraphics[width=\linewidth]{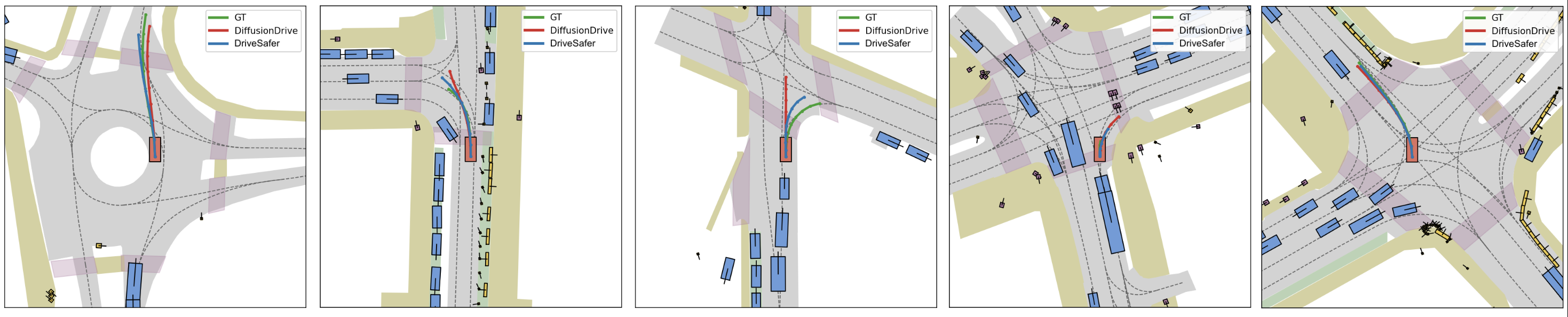}
    \vspace{-8mm}
    \caption{Sample instances where DiffusionDrive predictions had a PDMS score of 0 but were fixed with \textit{DriveSafer} for a safer trajectory.}
    \label{fig:improvements}
\vspace{-5mm}
\end{figure*}
During training, collision losses are calculated based on ground truth. For inference-time collision awareness, the motion of other road agents must be estimated well. We observe from Figure \ref{fig:diffdrive_error_dist} that many of the catastrophic failures for the baseline DiffusionDrive model involve collisions. However, DiffusionDrive only predicts agent locations for the current time step and the model is not optimized for forecasting the future locations of other agents. Hence, we add a lightweight adapter layer to predict the motion of other agents for future timesteps, enabling joint prediction and planning.
\subsection{Inference-Time Safety Guidance}
Once the model is trained with safety-aware losses, it will ideally learn to better avoid generating motion plans that violate safety. However, due to the stochastic nature of diffusion-based models, in addition to limitations in modeling sensor noise and complex scenes with many agents, enforcing safety at inference time requires attention.

We start inference-time safety guidance with alternative hypothesis generation. Since a motion plan for 4 seconds can have the ego vehicle moving as much as 100 m at highway speeds, an unsafe motion plan can potentially be made safe with a slight nudge in heading/progress which results in a relatively minor deviation of say 1 m over this time horizon. Motivated by this idea, we generate alternate trajectories that have a variation of $5\%$ in heading, speed and sub-meter lateral trajectory shifts to the top-most predicted mode from the diffusion model. Furthermore, we also use the second- and third-highest ranked modes of the diffusion model as alternate hypotheses. Once these are generated, we use the predicted positions of agents and the underlying map to assign a score to these alternative trajectories using safety metrics. The safest trajectory is then chosen as the final output, enabling safety guidance during inference. 

Note that we assume ego-vehicle localization is correct to sub-lane level accuracy, to be able to use the map information, but this is a reasonable assumption for many modern AVs. Visual localization methods such as MapLocNet \cite{wu2024maplocnet} and BEVMapMatch \cite{sural2026bevmapmatchmultimodalbevneural} have shown that such localization accuracy can be obtained reliably over time using standard datasets such as NuScenes even when GNSS fails. 

\section{Experiments and Results}

We use NAVSIM \cite{dauner2024navsim} as our benchmark for evaluating planner outputs due to its variety of complex scenarios in dense urban traffic such as the left turn shown in Figure \ref{fig:left_turn} which requires accurate planning. 
\begin{figure}
    \centering
    \includegraphics[width=\linewidth]{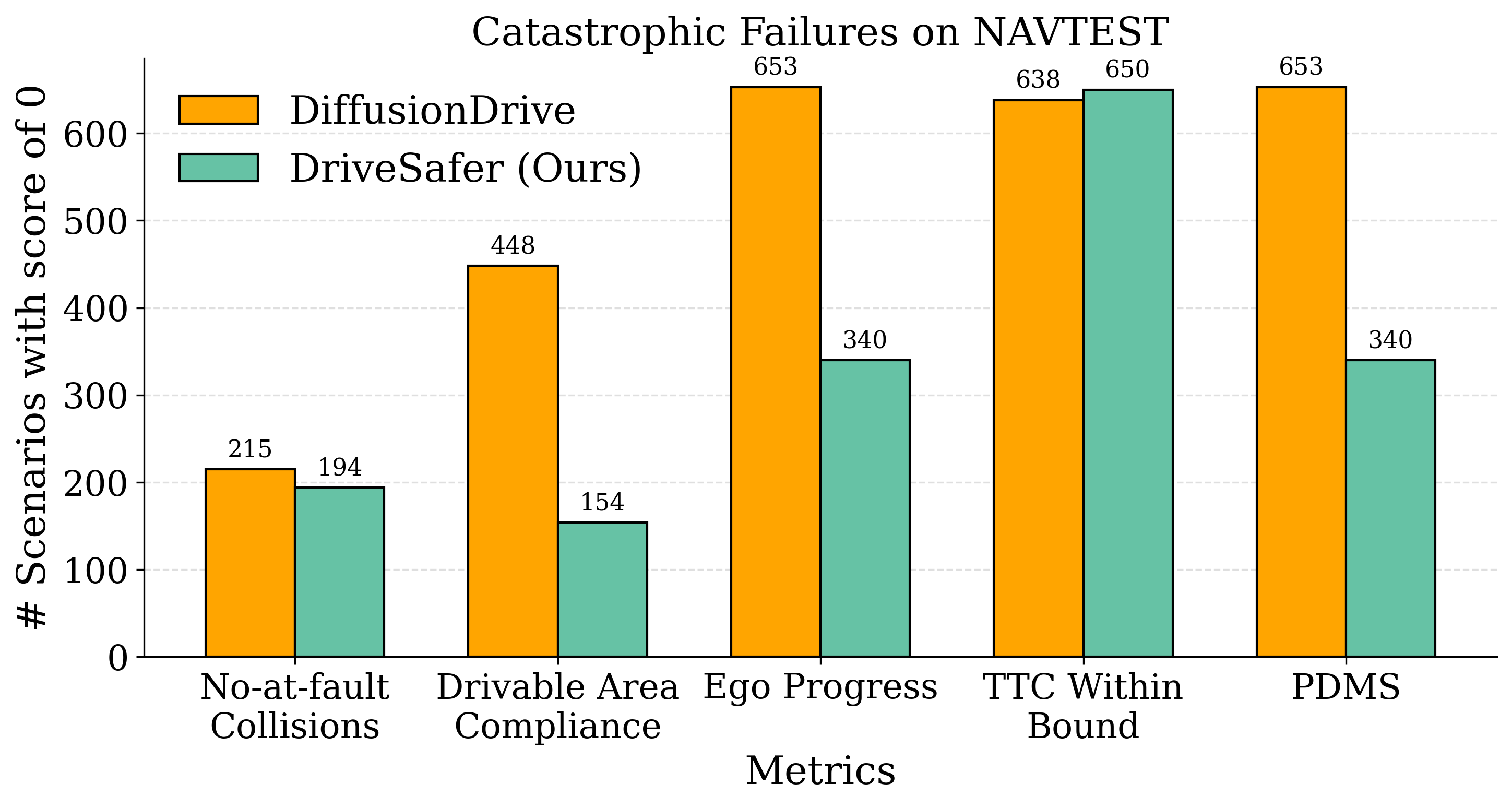}
    \vspace{-8mm}
    \caption{Analysis of catastrophic failure cases (PDM Score = 0) on NAVSIM navtest split. Lower values are better.}
    \label{fig:catastrophic_failure_plot}
\vspace{-6mm}
\end{figure}
Table \ref{tab:e2e_comparison} shows a comparative analysis of \textit{DriveSafer} against other state-of-the-art E2E planning methods on the NAVSIM benchmark. While our focus is on improving catastrophic failures, our model also improves the PDMS score by 2.2 points upon the primary SOTA baseline model DiffusionDrive pushing the metric to 90.3. Moreover, we see marked improvements in the Drivable Area Compliance (DAC) and Ego Progress (EP) sub-metrics. These improvements can be attributed to the use of the underlying map for enforcing safety and generating alternative trajectories when the primary trajectory violates drivable area requirements. Additionally, training with safety-aware losses also improves these metrics.

Figure \ref{fig:catastrophic_failure_plot} provides a more in-depth analysis of the catastrophic failure cases of DiffusionDrive which have \texttt{PDMS=0}. We see a marked improvement with the number of cases obtaining \texttt{PDMS=0} being reduced from 653 to 340 ($\approx48\%$). Moreover, on Drivable Area Compliance (DAC), the improvement is even better with only 154 cases with a DAC of 0 using \textit{DriveSafer}, as compared to 448 using DiffusionDrive, showing over $65\%$ reduction in this category of catastrophic failures.   

Among the 313 samples out of 12146 in \texttt{navtest} that yield an increase of PDMS score from 0 to a positive value, a few are illustrated in Figure \ref{fig:improvements}. The first instance shows the ground truth ego center trajectory in green and the trajectory produced by DiffusionDrive in red. The red planned trajectory results in the vehicle driving off the paved region, which is unsafe behavior. With \textit{DriveSafer}, an alternate safe path, shown in green, is found. The next two samples show a violation of the intended direction of travel with DiffusionDrive, which also gets corrected with \textit{DriveSafer}. The fourth sample shows a DiffusionDrive-generated path that collides with crossing pedestrians, but \textit{DriveSafer} stops the vehicle early avoiding the collision. In the last example, the red trajectory comes too close to an oncoming vehicle resulting in a collision, but \textit{DriveSafer} steers the trajectory away from such unsafe behavior with a slight nudge in the correct direction.
 
\section{Conclusion}
End-to-end autonomous driving models can be powerful, but their practicality is limited by the presence of a significant number of catastrophic failures that limit their real-world usability. We have presented \textit{DriveSafer}, a framework to systematically combat such failures. \textit{DriveSafer} deals with catastrophic failures through a combination of safety-aware losses during training and safety-guided inference. \textit{DriveSafer} reduces the number of catastrophic failures (\texttt{PDMS=0}) on the NAVSIM benchmark by $48\%$, obtaining a PDMS score of $90.3$ on the NAVSIM benchmark, outperforming existing state-of-the-art baselines on these metrics. More broadly, our results show that explicit safety guidance in E2E autonomous driving should be geared towards the direct mitigation of catastrophic failure cases. In the future, we will extend \textit{DriveSafer} for real-world deployment on an AV and assess its impact on safety in the real world, aiming to eventually bring catastrophic failures down to zero.
{
    \small
    \bibliographystyle{ieeenat_fullname}
    \bibliography{main}
}

\end{document}